# A novel teleoperator testbed to understand the effects of master-slave dynamics on embodiment and kinesthetic perception *

Mohit Singhala [1] *Student Member, IEEE* and Jeremy D. Brown[1] *Member, IEEE*

*Abstract*— With the rising popularity of telerobotic systems, the focus on transparency with regards to haptic perception is also increasing. Transparency, however, represents a theoretical ideal as most bilateral force-reflecting telerobots introduce dynamics (stiffness and damping) between the operator and the environment. To achieve true dexterity, it will be essential to understand how humans embody the dynamics of these telerobots and thereby distinguish them from the environment they are exploring. In this short manuscript, we introduce a novel single degree-of-freedom testbed designed to perform psychophysical and task performance assessments of kinesthetic perception during telerobotic exploration. The system is capable of being configured as a rigid mechanical teleoperator, a dynamic mechanical teleoperator, and an electromechanicaal teleoperator. We performed prefatory system identification and found that the system is capable of simulating telerobotic exploration necessary to understand the impact of master-slave dynamics on kinesthetic perception.

## I. INTRODUCTION

The human body possesses highly dexterous capabilities that are developed over the course of many years and refined with repeated practice [1]. While a large portion of our dexterous interactions involve the hand directly, there is substantial literature demonstrating our dexterous capabilities exhibited through a tool [2], [3]. One tool in particular, that has seen rapid adoption since its development, is the teleoperator. The advantage of a teleoperator is that is can extend the operator's manipulation abilities to environments that are not accessible through direct manipulation. As with the natural limbs [4], [5], dextrous manipulation through a teleoperator is only possible if the device's interactions with the environment are displayed to the operator in a manner consonant with their expectations.

With rare exception, most modern day teleoperators are robotic devices that can be found in applications ranging from space exploration to minimally invasive surgery. While these telerobots almost always provide the operator with visual feedback of the environment, haptic feedback, in particular direct force reflection is largely absent. This is due to the fact that the control architectures employed for most bilateral force-reflecting telerobots introduce dynamics (stiffness and damping) between the master and slave terminals that result in a tradeoff between stability of the telerobotic controller, and its overall performance [6]. Given that it can be inferred from dexterous tool use that humans are capable of incorporating tools with some inherent dynamics into their body schema, it is worth asking what impact these master-slave dynamics have on the operator's perception of the remote environment and separately, task performance in that environment.

In this short manuscript, we present a custom 1-DoF teleoperator testbed that is capable of rendering various mechanical and electromechanical transmissions between a given teleoperator master and slave. This reconfigurable teleoperator is designed to investigate how the master-slave dynamics of a teleoperator affect perception of the remote environment and to what extent these dynamics affect teleoperator embodiment and task performance. We believe that a perceptual understanding of the effects of teleoperators will be essential to achieve truly dexterous telerobotic manipulation.

## II. METHODS

The following sections provide an overview of the testbed design, sensing and control, and methods used for prefatory system analysis.

### A. Testbed Design

The testbed consists of a single degree of freedom kinesthetic feedback interface with three distinct modules shown in Fig. 1 and described below.

*1) Hand Fixture (Participant):* The first module serves as the input interface between the participant and the teleoperator testbed. Custom 3D-printed fixtures are used to enable different types of hand configurations to be used for haptic exploration. The hand fixture (see Fig. 1) is designed for an alternating finger grip, enabling exploration via pronation and supination of the forearm. The system can also be configured to use conventional handle grips for exploration along any of the three primary axes of wrist-rotation.

*2) Teleoperator:* The teleoperator comprises the three distinct transmissions described below. All transmissions are connected via capstan drives and rotate in the same direction as the hand fixture. Each transmission can be independently engaged and disengaged through shaft couplers.

- Rigid mechanical transmission: This transmission uses a rigid a stainless steel rod (8mm diameter) between the master and slave to couple the input from the hand to the environment with no torque or position scaling.
- Dynamic mechanical transmission: This transmission uses a torsional spring and rotary damper in parallel between the master and slave to couple the input from the hand to the environment. Input and output shafts from both the spring and the damper can be engaged and

*This material is based upon work supported by the National Science Foundation under NSF Grant #1910939

[1]Mohit Singhala and Jeremy D. Brown are with the Department of Mechanical Engineering, Johns Hopkins University, Baltimore, MD, USA. `mohit.singhala@jhu.edu`

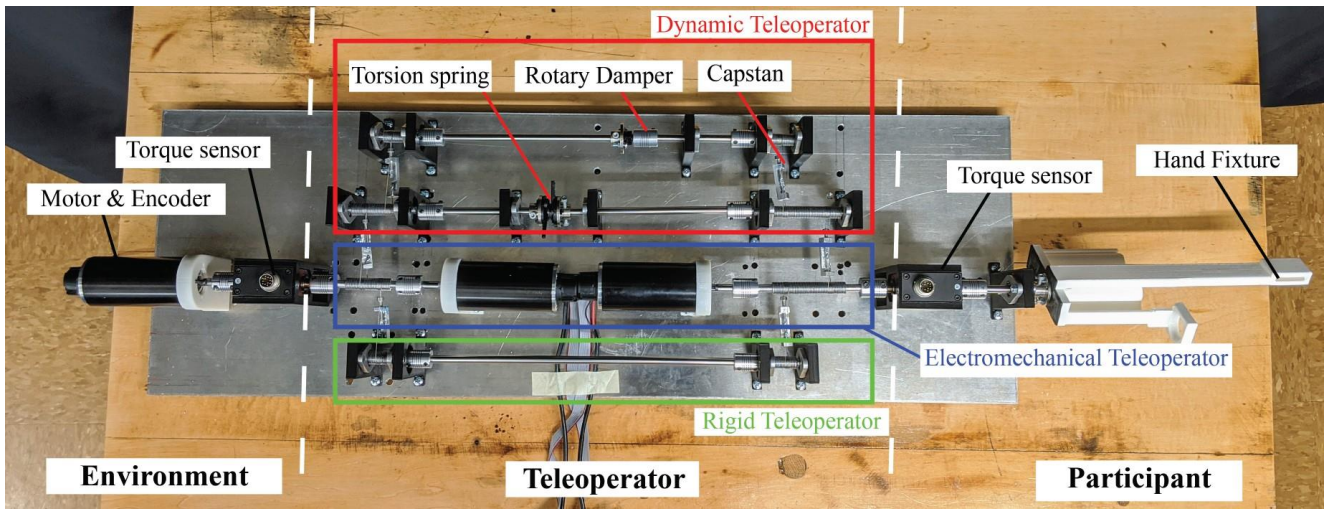

Fig. 1. Experimental setup with the three modules: 1) Participant- custom 3D printed hand fixture and torque sensor, 2) Teleoprator including three transmissions - rigid mechanical (green), electromechanical (blue), and dynamic mechanical (red), and 3) Environment - DC motor and encoder for virtual environment rendering and torque sensor.

disengaged independently, allowing this transmission to be configured as pure damping, pure stiffness or a combination of both. Currently, this transmission is configured to provide no torque or position scaling.
- Electromechanical transmission: This transmission uses two back-drivable Maxon RE50 (200 W) motors, each fit with a 3-channel 500 CPT HEDL encoder, as the master and slave. Using standard proportional-derivative control between the master and slave to couple the input from hand to the environment creates an impedance controlled electromechanical teleoperator with inherent dynamics (stiffness and damping).

The input and output shafts from the Electromechanical teleoperator act as the primary drive shafts and are connected to the Hand Fixture (input) and the Environment (output) respectively.

*3) Environment:* The environment is rendered at the free end of the primary output drive shaft. The output can be coupled to real environments or virtual environments rendered through the Maxon RE50 (200 W) motor and 500 CPT HEDL encoder shown in Fig. 1.

### B. Data Acquisition and Control

The motors are controlled using a Quanser AMPAQ L4 linear current amplifier. Two Futek non-contact rotary torque sensors (TRS600) with a 5 Nm maximum torque capacity are coupled to the input and output shafts of the teleoperator. These sensors measure the torque applied by the participant and the torque (input) rendered by the environment (output), respectively. All data acquisition and control operations are performed using a Quanser QPIDe DAQ, run at a 1Khz frequency via a MATLAB/Simulink and QUARC interface. A monitor is used for visual stimuli, and a pair of Bose noise cancelling-headphones provide the experimenter and participant with identical audio cues. The software interface is configured to run popular psychophysical paradigms including Methods of Constant Stimuli and adaptive staircases.

### C. Prefatory System Analysis

Matlab 2018b was used to analyze the rigid mechanical transmission with a virtual torsion spring (4mNm/deg) as the environment. The rendered system (Environment) and the system felt by the participant (Participant) were modeled as second order systems using the System Identification Toolbox. Input displacement and torque applied by the participant were used to identify the model for the Participant system and Output displacement and torque rendered by the virtual environment were used to identify the model for the Environment system. Results for this rigid teleoperator configuration are discussed in the following section.

## III. RESULTS

The step response and Bode plot of the Environment and Participant systems with the virtual torsion spring are overlaid for the rigid mechanical transmission in Fig. 3. The step response from the Participant system with no environment rendering (free space) is shown in Fig. 4 The goodness of fit statistics - percent fit, final prediction error (FPE), and mean-square error (MSE) - for each model are reported in Table I.

TABLE I
GOODNESS OF FIT STATISTICS

| System | Percent Fit | FPE | MSE |
|---|---|---|---|
| Environment | 99.94% | 3.006e-05 | 3.005e-05 |
| Participant | 99.95% | 2.716e-05 | 2.715e-05 |
| Free Space | 99.70% | 3.331e-03 | 3.309e-03 |

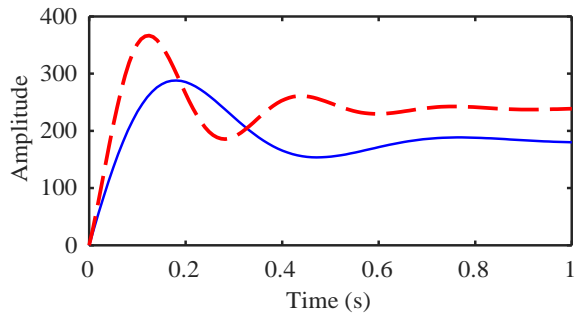

Fig. 2. Step response of the Environment (red-dashed) and Participant (blue-solid) systems when exploring a virtual torsional spring with the rigid mechanical teleoperator transmission.

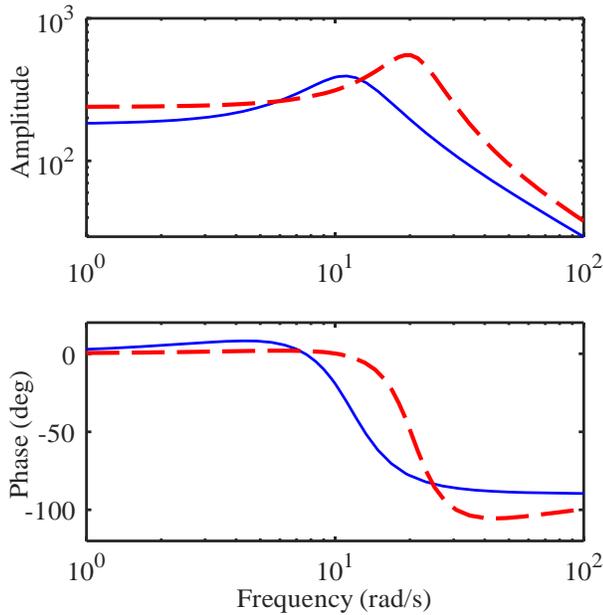

Fig. 3. Bode plot of the Environment (red-dashed) and Participant (blue-solid) systems when exploring a virtual torsional spring with the rigid mechanical teleoperator transmission.

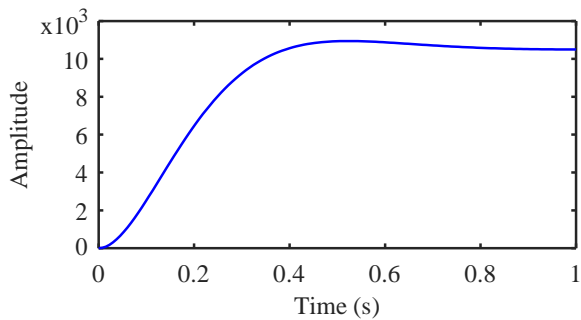

Fig. 4. Step response illustrating the inherent damping the participant feels while using the rigid mechanical teleoperator transmission to explore free space

## IV. DISCUSSION

The system response is in agreement with the response expected from a torsional spring dampened by the transmission. The red trace represents the environment rendered by the testbed, which includes the torsion spring rendered by the motor and viscous damping from the motor action and other losses due to friction. The blue trace represents the system felt by the participant and includes viscous damping from the transmission, and is therefore, more damped than the rendered environment- as evident from step response of the system when no virtual environment is rendered (See Fig. 2). The differences in the two traces represents the master-slave dynamics that we wish to study through this testbed.

The system response here only reflects the testbed's potential to replicate common telerobotic exploration that has been studied extensively in literature [7], [8]. We plan to perform psychophysical assessments of participants for different telerobotic configurations for the same virtual environments to gain a perceptual understanding of teleoperator embodiment. We will supplement this with performance tasks that can reflect this understanding in practical telerobotic use cases through different virtual and real environments.

## V. CONCLUSIONS

We presented a 1-DoF teleoperator testbed capable of rendering virtual and mixed reality haptic environments with three different transmissions- rigid mechanical, dynamic mechanical, and electromechanical. We performed system identification of the rigid mechanical transmission for the Participant and Environment systems, while rendering a virtual torsion spring. The results show a robust estimate for the master-slave dynamics and validate the testbed as a platform to research perceptual embodiment of teleoperators. An in-depth analysis of all the configurations of the system is planned before the testbed is used for human-subjects research.

## ACKNOWLEDGMENT

We would like to thank Evan Pezent (Rice University) for his assistance with the design of the capstan drives.